\documentclass[letterpaper, 10 pt, conference]{ieeeconf}  
\usepackage{cite}
\usepackage{amsmath,amssymb,amsfonts}
\usepackage{algorithmic}
\usepackage{graphicx}
\usepackage{textcomp}
\usepackage{xcolor}
\usepackage{CJKutf8}
\usepackage[switch]{lineno}
\usepackage{multirow}
\usepackage{array}
\usepackage{bm}
\usepackage{float}
\usepackage{booktabs}

\title{\LARGE \bf
ViewMind3D: Modular View-Aware Inference for Training-Free 3D-QA}

\author{Ping-Kun Chiang, Kun-Ru Wu, Po-han Li, Sandeep Chinchali, Ufuk Topcu, Yu-Chee Tseng}

\begin{document}

\maketitle
\thispagestyle{empty}
\pagestyle{empty}

%%%%%%%%%%%%%%
\begin{abstract}
Recent advances in large language models (LLMs) and vision-language models (VLMs) have enabled new possibilities for 3D question answering (3D-QA), a key capability for embodied AI and robotic perception. 
However, most existing methods rely on 3D-specific training or fine-tuning with costly annotations, limiting their scalability and real-world applicability.
We present \textbf{ViewMind3D}, a fully training-free and modular framework for 3D spatial reasoning over multi-view observations of a scene without requiring complete 3D reconstruction. 
The framework decomposes the 3D-QA task into four interpretable components: (1) question-driven multi-view selection, (2) guided visual grounding with language-conditioned object cues, (3) spatial context encoding via a bird's-eye-view (BEV) viewpoint indicator, and (4) structured answer generation through role-based reasoning. 
This design enables structured, robust, and interpretable reasoning without requiring model tuning.
Experimental results on ScanQA and SQA3D show that ViewMind3D achieves competitive performance compared to prior training-free and fine-tuned 3D-LLMs. 
In particular, our method improves performance on spatially grounded question types, such as ``What'' questions in SQA3D, while maintaining strong overall accuracy (50.8\%) and achieving 73.4 CIDEr on ScanQA.
These results demonstrate that effective 3D reasoning can be achieved through modular orchestration of general-purpose LLMs and VLMs for robotic perception in real-world environments.
\end{abstract}

%%%%%%%%%%
\section{Introduction}

Understanding 3D scenes from language queries is a core capability for embodied AI, robotics, and human--robot interaction.
While recent advances in 3D question answering (3D-QA) improve spatial reasoning, most methods rely on costly 3D annotations, task-specific training, and specialized architectures~\cite{hong20233dllm,xu2024pointllm,qi2024shapellm}, limiting scalability.
In contrast, vision--language models (VLMs)~\cite{openai2023gpt4v} generalize well but lack explicit 3D reasoning mechanisms.
This raises a key question: {\em Can general-purpose 2D VLMs perform 3D reasoning without retraining or fine-tuning?}

A 3D-QA system connects spatial language understanding with grounded visual reasoning. 
Given a 3D scene represented by multi-view observations, a question, and optionally the questioner's state, the goal is to produce a grounded answer. 
Benchmarks such as ScanQA~\cite{azuma2022scanqa} and SQA3D~\cite{ma2022sqa3d} evaluate 3D vision--language reasoning in complex indoor environments. 
Existing methods span 2D-based, 3D-based, and hybrid paradigms~\cite{hong20233dllm,xu2024pointllm,qi2024shapellm,chen2024grounded}, but typically require retraining and remain dataset-specific, limiting generalization across environments.

By integrating with large language models (LLMs), \cite{SceneLLM} leverages large-scale pretraining and language--3D alignment, while \cite{GPT4Scene} employs visual prompting to capture global--local relationships. 
Object-centric approaches such as \cite{ChatScene, ChatScene++} further enable fine-grained grounding and compositional reasoning through structured object representations and interactions. 
However, these methods typically rely on substantial 3D-aware training, and their performance often degrades under strict zero-shot settings. Consequently, they depend on training data, prompt design, or task-specific tuning, which limits their generalization to unseen environments.

Most existing 3D-QA methods therefore face key challenges for robotics applications, including high annotation cost, computational overhead, and limited reasoning capability of smaller backbones~\cite{hong20233dllm}. Although recent efforts leverage pre-trained LLMs and VLMs for training-free reasoning~\cite{hong20233dllm,yang2024llmgrounder}, they often still depend on auxiliary 3D-trained components, leaving fully modular, interpretable, and training-free solutions underexplored.

To address these limitations, we propose \textbf{ViewMind3D}, a modular, training-free framework for 3D-QA in real-world indoor scenes. 
With a structured design, it performs 3D-QA without task-specific fine-tuning or parameter updates, while supporting general-purpose prompts. 
The framework decomposes 3D reasoning into interpretable components, enabling general-purpose LLMs/VLMs to reason over multi-view observations, even in unseen environments. 
This design enables robust, interpretable 3D reasoning while remaining flexible across diverse LLMs/VLMs.

ViewMind3D consists of four components: (1) \textbf{Relevance View Selection}, filtering multi-view images for question-relevant views; 
(2) \textbf{Guided Detection}, overlaying visual annotations for referenced objects; 
(3) \textbf{Viewpoint Indicator}, encoding camera poses as bird's-eye-view (BEV) representations; and 
(4) \textbf{3D-QA Module}, fusing visual, spatial, and linguistic cues to generate grounded answers.

These components address key challenges in multi-view 3D reasoning. Multi-view inputs often contain redundant or irrelevant information, motivating relevance-based selection. The absence of explicit object grounding weakens alignment between visual evidence and language queries, which is mitigated by guided detection. Finally, individual views lack a shared spatial reference, hindering cross-view reasoning; this is addressed by viewpoint indicators encoded in BEV.

Experimental results show that ViewMind3D achieves competitive performance against prior methods, including 3D-LLM~\cite{hong20233dllm}, improving CIDEr by 4.01 on ScanQA and achieving higher overall accuracy on SQA3D in zero-shot settings, demonstrating the effectiveness of our modular design. The contributions of this work are:
\begin{itemize}
\item A modular and fully training-free 3D-QA framework enabling 2D VLMs to perform spatial reasoning over multi-view observations without retraining;
\item A question-driven visual grounding module providing phrase-level bounding box annotations that improve interpretability and reduce visual noise;
\item A BEV-based viewpoint encoding that represents camera position and orientation as interpretable spatial cues;
\item A structured multi-stage reasoning process that aggregates cross-view information into spatially grounded representations for robust 3D understanding.
\end{itemize}

\section{Related Works}

Text-only LLMs~\cite{Achiam2023GPT4TR} lack grounding in non-linguistic modalities. 
VLMs, exemplified by CLIP and its successors~\cite{Radford2021CLIP,Li2022BLIPBL}, align visual and textual representations through large-scale pretraining. 
Building on this foundation, modern VLMs integrate visual encoders with LLMs via lightweight adapters~\cite{Li2023BLIP2BL,Liu2023VisualIT}, enabling cross-modal reasoning over images and language. 
Recent large multimodal models~\cite{openai2023gpt4v} further extend these capabilities to support multi-step reasoning and zero-shot generalization across diverse perception tasks. Despite these advances, existing VLMs primarily operate on 2D observations and lack explicit mechanisms for reasoning over 3D spatial structures.

3D vision--language reasoning integrates spatial, visual, and linguistic information, enabling agents to understand complex 3D environments. 
Benchmarks such as ScanQA~\cite{azuma2022scanqa} and SQA3D~\cite{ma2022sqa3d} evaluate tasks including 3D-QA, object grounding, and scene understanding. 
Existing approaches can be categorized as:
\begin{itemize}
\item 
\textbf{2D-based methods}~\cite{hong20233dllm}: project 3D scenes into multi-view images or videos, leveraging strong 2D priors but lacking explicit spatial modeling.
\item 
\textbf{3D-based methods}~\cite{xu2024pointllm,qi2024shapellm}: directly process point clouds or meshes to capture geometric structure, requiring specialized architectures and training.
\item 
\textbf{Hybrid methods}~\cite{chen2024grounded,chen2024ll3da}: combine 2D and 3D representations for unified scene understanding, often at the cost of increased model complexity.
\end{itemize}

Despite recent progress, prior methods still face key challenges, including high annotation cost, computational overhead, and limited generalization due to reliance on training-specific components~\cite{hong20233dllm}. 
Scene-LLM~\cite{SceneLLM} improves spatial understanding through large-scale pretraining and alignment between language and 3D representations, enabling reasoning over structured scenes. However, Scene-LLM depends on large-scale 3D-language pretraining and curating substantial 3D-aware training data requires significant efforts.
GPT4Scene~\cite{GPT4Scene} adopts a visual prompting paradigm that augments multi-view inputs with BEV representations and spatial markers to establish global--local correspondence. However, its effectiveness in purely training-free settings remains limited.
Incorporating multi-view information into 3D datasets has also been explored in DSPNet~\cite{DSPNet} and MV-ScanQA~\cite{MVScanQA}, demonstrating strong performance. However, these approaches rely on training or pre-training for 3D feature fusion.
Object-centric approaches such as Chat-Scene~\cite{ChatScene} and Chat-Scene++~\cite{ChatScene++} represent scenes as sequences of object-level embeddings with identifier tokens, enabling fine-grained grounding and reasoning. 
However, these methods rely on learned object representations and scene-language training.
Further, these methods require additional adaptation and often entangle perception and reasoning into monolithic pipelines, reducing interpretability.

While recent works explore zero-shot strategies using pre-trained LLMs and VLMs~\cite{hong20233dllm,yang2024llmgrounder}, most still depend on auxiliary 3D-trained modules, limiting scalability and interpretability. In contrast, ViewMind3D adopts a fully training-free and modular approach, enhancing interpretability.

\section{Methodology}

\begin{figure*}[t]
    \centering
    \vspace{10pt}
    \includegraphics[width=0.85\linewidth]{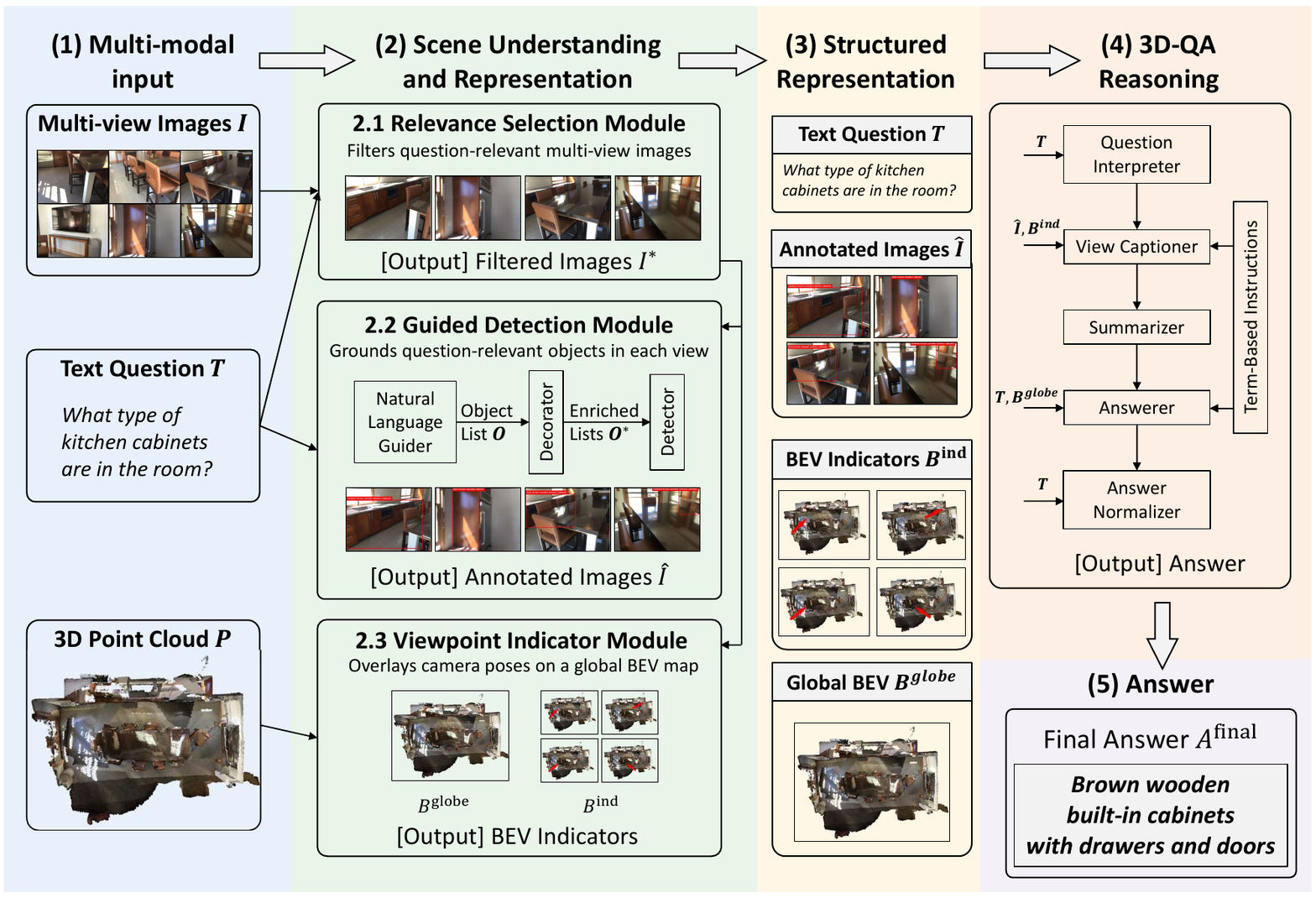}
\caption{Overview of the ViewMind3D framework for 3D-QA. 
Given a natural language query, multi-view images, and a 3D point cloud, the system performs scene understanding and organizes the outputs into a structured representation that bridges perception and reasoning. 
This representation is then processed through four stages. 
The example illustrates how the query is grounded in multi-view visual evidence and resolved via structured spatial reasoning.}

    \label{fig:framework}
\end{figure*}

\subsection{Problem Statement}

Given a 3D scene $\mathcal{S}$ represented by a point cloud $\bm{P} \in \mathbb{R}^{N_{\mathrm{pts}} \times D}$, where $N_{\mathrm{pts}}$ is the number of points and $D$ denotes the feature dimensionality (e.g., coordinates, color, or semantics), the goal of 3D question answering (3D-QA) is to generate a textual answer $\mathcal{A}$ for a natural language query $T$. All points in $\bm{P}$ are defined in a unified global coordinate frame.

Optionally, the input includes (i) a situation descriptor $s = \langle s^{\mathrm{txt}}, s^{\mathrm{pos}}, s^{\mathrm{ori}} \rangle$ and (ii) a set of multi-view RGB images $\bm{I} = \{ I_1, \dots, I_{N_{\mathrm{view}}} \}$, defined from the perspective of the questioner. 
Here, $s^{\mathrm{txt}}$ describes the spatial context, $s^{\mathrm{pos}} \in \mathbb{R}^3$ denotes the 3D position, and $s^{\mathrm{ori}} \in SO(3)$ represents the orientation. 
SQA3D~\cite{ma2022sqa3d} provides $s$, while ScanQA~\cite{azuma2022scanqa} does not. 
Both datasets are built on ScanNet~\cite{dai2017scannet}. 
Each image $I_i \in \mathbb{R}^{H \times W \times 3}$.

The 3D-QA task is defined as a mapping
${F} : \left( \bm{P}, T, s, \bm{I} \right) \mapsto \mathcal{A}$,
where $s$ and $\bm{I}$ are optional. 
The objective is to produce an answer $\mathcal{A}$ that is semantically consistent with $T$ and grounded in $\mathcal{S}$.

In this work, $F$ is realized in a fully training-free manner, without task-specific fine-tuning or parameter updates on 3D data, using pre-trained LLMs and VLMs.

\subsection{ViewMind3D Framework}

ViewMind3D is a modular, fully training-free framework for open-ended 3D-QA. 
As illustrated in Fig.~\ref{fig:framework}, it consists of four stages: relevance selection, visual grounding, spatial encoding, and answer synthesis.

We introduce an intermediate \textbf{structured representation} that bridges perception and reasoning. 
Specifically, outputs from scene understanding modules are organized into filtered views, grounded visual annotations, and spatial BEV cues, enabling modular and interpretable 3D reasoning.

\textbf{Relevance View Selection:}
Filters multi-view observations to retain question-relevant perspectives while preserving high recall for downstream reasoning.

\textbf{Guided Detection:}
Extracts object cues from $T$ and grounds them in selected views via open-vocabulary detection, producing annotated images for visual grounding.

\textbf{Viewpoint Indicator:}
Encodes each view with a bird's-eye-view (BEV) representation of camera position and orientation, providing explicit spatial context.

\textbf{3D-QA Module:}
Given the structured representation, integrates visual, spatial, and linguistic information through multi-step reasoning to generate grounded answers.

All components operate in an inference-only regime without parameter updates. 
The framework leverages general-purpose LLMs and VLMs (e.g., GPT-4.1 and o3), enabling a fully training-free yet effective 3D-QA pipeline.

\subsection{Relevance View Selection}
\label{sec:method_rsm}

Given multi-view images $\bm{I}$ and a query $T$, this module filters views relevant for answering $T$. 
For each image $I_i \in \bm{I}$, a relevance-selection agent predicts a binary label:
\begin{equation}
r_i = 
\begin{cases}
  \mathrm{Agent}_{\mathrm{selector}}(T, I_i), & \text{if ScanQA,} \\
  \mathrm{Agent}_{\mathrm{selector}}(T, s^{\mathrm{txt}}, I_i), & \text{if SQA3D,}
\end{cases}
\end{equation}
where $r_i \in \{\mathrm{yes},\, \mathrm{no}\}$ indicates whether the image is retained.

The filtered set is $\bm{I}^* = \{ I_i \mid r_i = \mathrm{yes} \}$ with $N^*_{\mathrm{view}} = |\bm{I}^*|$. 
Instead of enforcing a fixed number of views, the module adopts a query-adaptive strategy that prioritizes recall, preserving potentially useful spatial evidence.

Multi-view inputs often contain redundant observations that hinder reasoning. 
By selecting question-relevant views, this module reduces redundancy and provides a more focused input for subsequent processing, improving the stability of downstream reasoning without requiring additional training.

\subsection{Guided Detection}

This module aligns question intent with visual content by transforming filtered views $\bm{I}^*$ into annotated images $\bm{\hat I}$ that highlight question-relevant objects. 
These annotations form part of structured representation for downstream reasoning.

\subsubsection{Natural Language Guider}
\label{sec:method_gdm_natural_language_guider}

A language-guided agent extracts object categories from the query:
\begin{equation}
\bm{O} =
\begin{cases}
\mathrm{Agent}_{\mathrm{guider}}(T), & \text{if ScanQA,} \\
\mathrm{Agent}_{\mathrm{guider}}(T, s^{\mathrm{txt}}), & \text{if SQA3D,}
\end{cases}
\end{equation}
where $N_\mathrm{obj} = |\bm{O}|$. 
If no object is referenced, grounding is skipped. 
For example, for ``What type of kitchen cabinets are in the room?'', the output is $\bm{O} = \{\text{kitchen cabinets}\}$.

\subsubsection{Decorator}
\label{sec:method_gdm_decorator}

For each image $I^{*}_i \in \bm{I}^{*}$, a decorator agent enriches each object $o_j \in \bm{O}$ with attributes:
\begin{equation}
\bm{O}^{*}_i = \mathrm{Agent}_{\mathrm{decorator}}(\bm{O}, I^{*}_i).
\end{equation}
This produces image-specific object phrases $\bm{O}^{*} = \{ \bm{O}^{*}_1, \ldots, \bm{O}^{*}_{N^{*}_{\mathrm{view}}} \}$, providing more discriminative cues for grounding.

\subsubsection{Detector}
\label{sec:method_gdm_detector}

An object detection agent localizes objects in $\bm{O}^{*}_i$ using Florence-2~\cite{xiao2024florence2}. 
For each $I^{*}_i \in \bm{I}^{*}$:
\begin{equation}
\bm{A}_i = \{ (b_{i,k}, o^i_k) \mid o^i_k \in \bm{O}^{*}_i \},
\end{equation}
where $b_{i,k}$ denotes the bounding box of $o^i_k$. 
Non-maximum suppression retains boxes with IoU below $\tau = 0.6$, yielding $\hat{\bm{A}}_i = \mathrm{NMS}(\bm{A}_i, \tau = 0.6)$. 
The annotated image is $\hat{I}_i = \mathrm{Overlay}(I^{*}_i, \hat{\bm{A}}_i)$, producing $\hat{\bm{I}} = \{ \hat{I}_1, \ldots, \hat{I}_{N^{*}_{\mathrm{view}}} \}$.

If no objects are detected, the image remains unchanged. 
For ``How''-type questions in SQA3D, annotations are omitted to avoid unreliable detections. 
These annotations provide explicit visual grounding for subsequent reasoning.

\subsection{Viewpoint Indicator}

This module encodes camera poses as explicit spatial cues using bird's-eye-view (BEV) representations. 
It generates a global BEV $B^{\mathrm{globe}}$ from the 3D scene $\mathcal{S}$ and $N^{*}_{\mathrm{view}}$ view-specific indicator maps for each $I^{*}_i \in \bm{I}^{*}$.

Each camera pose $C_i \in SE(3)$, defined by rotation $R_i \in \mathbb{R}^{3\times3}$ and translation $T_i \in \mathbb{R}^3$, is projected onto the BEV plane. 
The indicator map is constructed as
\begin{equation}
B^{\mathrm{ind}}_i = \mathrm{Overlay}(B^{\mathrm{globe}}, C_i),
\end{equation}
where $\mathrm{Overlay}(\cdot)$ draws a directional marker at $T_i$ based on $R_i$. 
The full set is $\bm{B}^{\mathrm{ind}} = \{ B^{\mathrm{ind}}_1, \ldots, B^{\mathrm{ind}}_{N^*_{\mathrm{view}}} \}$.

These BEV indicators provide a shared spatial reference across views, enabling consistent reasoning over viewpoint relationships and scene layout. 
They form part of the structured representation used for downstream reasoning.

\subsection{3D-QA Module}
\label{sec:3dqa_module}

The 3D-QA module performs structured multimodal reasoning based on the structured representation, including the question $T$, annotated images $\bm{\hat{I}}$, global BEV $B^{\mathrm{globe}}$, and viewpoint indicators $\bm{B}^{\mathrm{ind}}$. 
As shown in Fig.~\ref{fig:framework}, it consists of five stages: question interpretation, view captioning, cross-view summarization, answer synthesis, and normalization.

\subsubsection{Question Interpreter}
\label{sec:method_3dqa_question_interpreter}

The interpreter extracts structured intent from the input question:
\begin{equation}
\mathcal{Q} =
\begin{cases}
\mathrm{Agent}_{\mathrm{interpreter}}(T), & \text{if ScanQA,} \\
\mathrm{Agent}_{\mathrm{interpreter}}(T, s^{\mathrm{txt}}), & \text{if SQA3D.}
\end{cases}
\end{equation}
$\mathcal{Q} = (\mathcal{O}_q,\, \mathcal{R}_q,\, \mathcal{S}_q,\, \mathcal{F}_q)$ encodes entities, reasoning type, spatial constraints, and attribute filters, providing a structured guide for downstream reasoning.

\subsubsection{View Captioner}
\label{sec:method_3dqa_view_captioner}

For each view, a captioning agent generates a viewpoint-aware description:
\begin{equation}
\langle L^{\mathrm{view}}_i, S^{\mathrm{view}}_i \rangle = \mathrm{Agent}_{\mathrm{captioner}} \big( \mathcal{Q}, \hat{I}_i, B^{\mathrm{ind}}_i \big),
\end{equation}
where $L^{\mathrm{view}}_i$ encodes viewpoint information and $S^{\mathrm{view}}_i$ summarizes question-relevant observations. 
The outputs are $\bm{L}^{\mathrm{view}}$ and $\bm{S}^{\mathrm{view}}$. 
BEV indicators provide spatial cues for consistent cross-view interpretation.

\subsubsection{Summarizer}
\label{sec:method_3dqa_summarizer}

The summarizer aggregates view-level descriptions into compact region-level representations:
\begin{equation}
\begin{split}
\bm{M}^{\mathrm{region}} 
&= \mathrm{Agent}_{\mathrm{summarizer}}(\bm{L}^{\mathrm{view}}, \bm{S}^{\mathrm{view}}) \\
&= \{ M^{\mathrm{region}}_1, M^{\mathrm{region}}_2, \ldots, M^{\mathrm{region}}_{N_{\mathrm{region}}} \}.
\end{split}
\end{equation}
Each region summarizes spatially coherent information, reducing redundancy across views.

\subsubsection{Answerer}
\label{sec:method_3dqa_answerer}

The answerer generates the final response using region summaries and global spatial context:
\begin{equation}
\mathcal{A} = 
\begin{cases}
\mathrm{Agent}_{\mathrm{answerer}}(T, \bm{M}^{\mathrm{region}}, B^{\mathrm{globe}}), & \text{if ScanQA,} \\
\mathrm{Agent}_{\mathrm{answerer}}(T, s^{\mathrm{txt}}, \bm{M}^{\mathrm{region}}, B^{\mathrm{globe}}), & \text{if SQA3D.}
\end{cases}
\end{equation}

\subsubsection{Answer Normalizer}
\label{sec:method_3dqa_answer_normalizer}

For ScanQA, the output is normalized for evaluation:
\begin{equation}
\mathcal{A}^{\mathrm{final}} = 
\begin{cases}
\mathrm{Agent}_{\mathrm{normalizer}}(T, \mathcal{A}), & \text{if ScanQA}
\\
\mathcal{A}, & \text{if SQA3D}
\end{cases}
\end{equation}

\subsubsection{Term-Based Instructions}
\label{sec:method_3dqa_tbi}

This module constrains outputs to dataset-specific vocabularies, reducing variation and improving evaluation consistency. 
Prompt details are provided in the multimedia material.

\section{Experiments}

\subsection{Datasets}
\label{sec:exp_datasets}

We evaluate \textbf{ViewMind3D} on two standard 3D-QA benchmarks, \textbf{ScanQA}~\cite{azuma2022scanqa} and \textbf{SQA3D}~\cite{ma2022sqa3d}, both built on ScanNet~\cite{dai2017scannet}. 
SQA3D additionally provides a textual situation descriptor for each question.
Following prior works~\cite{azuma2022scanqa,ma2022sqa3d,hong20233dllm}, we evaluate on the ScanQA validation set and the SQA3D test set, ensuring consistent comparison with existing methods. 
All experiments are conducted in a zero-shot, training-free setting without any task-specific fine-tuning.
Multi-view RGB images are obtained from the $\mathrm{scannet\_frames\_25k}$ subset, and the corresponding 3D point clouds are used to construct BEV-based viewpoint indicators.

\begin{table*}[t]
\centering
\vspace{10pt}
\caption{Comparisons on the ScanQA validation set.}
\label{exp_scanqa_val_main_results}
\begin{tabular}{lcccccc}
\toprule
 & BLEU-1 $\uparrow$ & METEOR $\uparrow$ & ROUGE-L $\uparrow$ & CIDEr $\uparrow$ & Cos. Sim. $\uparrow$ & BERTScore-F1 $\uparrow$ \\
\midrule

VoteNet+MCAN                & 28.00 & 11.40 & 29.80  & 54.70 & -- & -- \\
ScanRefer+MCAN              & 26.90 & 11.50 & 30.00  & 55.40 & -- & -- \\
ScanQA                      & 30.20 & 13.10 & 33.30  & 64.90 & -- & -- \\

\midrule
flamingo-SingleImage        & 23.80 & 10.70 & 26.70  & 52.00 & -- & -- \\
flamingo-MultiView          & 25.60 & 11.30 & 31.10  & 55.00 & -- & -- \\
BLIP2-flant5-SingleImage    & 28.60 & 10.60 & 25.80  & 42.60 & -- & -- \\
BLIP2-flant5-MultiView      & 29.70 & 11.30 & 26.60  & 45.70 & -- & -- \\

\midrule
3D-LLM (flamingo)           & 30.30 & 12.20 & 32.30  & 59.20 & -- & -- \\
3D-LLM (BLIP2-opt)          & 35.90 & 13.80 & 34.00  & 63.80 & -- & -- \\
3D-LLM (BLIP2-flant5)       & \textbf{39.3} & 14.50 & 35.70 & 69.40 & -- & --\\

\midrule
LLaVA-SingleImage           &  7.10 & 10.50 & 12.30  & 5.70 & -- & -- \\
GPT-4.1-SingleImage         & 19.34 & 7.87 & 16.84 & 7.90 & 32.72 & 49.33 \\
GPT-4.1-MultiView           & 4.28 & 9.11 & 9.59 & 0.97 & 43.16 & 82.93 \\
GPT-4.1-TwoStage            & 10.92 & 13.61 & 25.68 & 35.97 & 51.67 & 77.42 \\
GPT4Scene (Qwen2-VL-7B, zero-shot) & -- & 14.10 & 33.2 & 68.7 & -- & -- \\
ViewMind3D (GPT-4.1)        & 34.33 & 14.95 & 35.61 & 68.30 & 62.29 & 89.78 \\
ViewMind3D (o3)             & 35.82 & \textbf{15.34} & \textbf{37.62} & \textbf{73.41} & \textbf{63.74} & \textbf{91.94} \\

\bottomrule
\end{tabular}
\end{table*}

\begin{table*}[t]
\centering
\caption{Comparisons on the SQA3D test set. \\
Format: V = 3D visual scene $\mathcal{S}$; 
S = situation description $s^{\mathrm{txt}}$; 
Q = question $T$; 
A = answer $\mathcal{A}$; 
L = location $\langle s^{\mathrm{pos}}, s^{\mathrm{ori}} \rangle$.}
\label{tab:sqa3d_main_result}

\begin{tabular}{lccccccccc}
\toprule
 & Format & What & Is & How & Can & Which & Others & Avg. \\
\midrule

ScanQA(w/o $s^{\mathrm{txt}}$) & VQ $\rightarrow$ A & 28.58 & 65.03 & 47.31 & 66.27 & 43.87 & 42.88 & 45.27 \\
ScanQA                         & VSQ $\rightarrow$ A & 31.64 & 63.80 & 46.02 & \textbf{69.53} & 43.87 & 45.34 & 46.58 \\
ScanQA + aux. task             & VSQ $\rightarrow$ AL & 33.48 & \textbf{66.10} & 42.37 & \textbf{69.53} & 43.02 & 46.40 & 47.20 \\

\midrule
MCAN                           & VSQ $\rightarrow$ A & 28.86 & 59.66 & 44.09 & 68.34 & 40.74 & 40.46 & 43.42 \\
ClipBERT                       & VSQ $\rightarrow$ A & 30.24 & 60.12 & 38.71 & 63.31 & 42.45 & 42.71 & 43.31 \\

\midrule
Unified QA$_{\mathrm{Large}}$ w/ ScanRefer & VSQ $\rightarrow$ A & 33.01 & 50.43 & 31.91 & 56.51 & 45.17 & 41.11 & 41.00 \\
Unified QA$_{\mathrm{Large}}$ w/ ReferIt3D & VSQ $\rightarrow$ A & 27.58 & 47.99 & 34.05 & 59.47 & 40.91 & 39.77 & 38.71 \\
GPT-3 w/ ScanRefer                          & VSQ $\rightarrow$ A & 39.67 & 45.99 & 40.47 & 45.56 & 36.08 & 38.42 & 41.00 \\
GPT-3 w/ ReferIt3D                          & VSQ $\rightarrow$ A & 28.90 & 46.42 & 28.05 & 40.24 & 30.11 & 36.07 & 34.57 \\
GPT-4.1-SingleImage                         & VSQ $\rightarrow$ A & 20.83 & 26.99 & 15.70 & 19.53 & 22.22 & 24.38 & 20.83 \\
GPT-4.1-MultiView                           & VSQ $\rightarrow$ A & 27.55 & 59.97 & 26.45 & 57.10 & 37.61 & 43.64 & 39.84 \\
GPT-4.1-TwoStage                            & VSQ $\rightarrow$ A & 23.80 & 47.09 & 30.54 & 34.02 & 44.16 & 30.57 & 33.11 \\
GPT4Scene (Qwen2-VL-7B, zero-shot)               & VSQ $\rightarrow$ A & -- & -- & -- & -- & -- & -- & 43.1 \\
GPT4Scene (GPT-4o, zero-shot)               & VSQ $\rightarrow$ A & -- & -- & -- & -- & -- & -- & 44.8 \\
GPT4Scene (Gemini-1.5-Pro, zero-shot)               & VSQ $\rightarrow$ A & -- & -- & -- & -- & -- & -- & 46.0 \\

\midrule
3D-LLM                                      & VSQ $\rightarrow$ A & 37.05 & 65.18 & 45.81 & 67.46 & \textbf{51.00} & 49.82 & 49.79 \\

\midrule
ViewMind3D (OpenAI o3)                      & VSQ $\rightarrow$ A & \textbf{44.38} & 64.11 & \textbf{47.53} & 59.47 & 37.89 & \textbf{53.71} & \textbf{50.75} \\

\bottomrule
\end{tabular}
\end{table*}

\subsection{Experimental Setup}

We compare ViewMind3D with both fine-tuned and zero-shot baselines. 
Fine-tuned methods include representative 3D approaches (e.g., VoteNet+MCAN, ScanRefer+MCAN, ScanQA~\cite{chen2020scanrefer,azuma2022scanqa}) and vision--language models such as Flamingo and BLIP2-FlanT5~\cite{alayrac2022flamingo,Li2023BLIP2BL}, as well as 3D-LLM~\cite{hong20233dllm}. 
Zero-shot baselines include LLaVA~\cite{Liu2023VisualIT} and GPT-4.1 variants ({\em SingleImage}, {\em MultiView}, {\em TwoStage})~\cite{Achiam2023GPT4TR}.
We also include GPT4Scene~\cite{GPT4Scene} for reference, reporting its zero-shot results for fair comparison with training-free methods, as its full performance relies on additional fine-tuning.

We report standard language generation metrics, including BLEU-1, METEOR, ROUGE-L, CIDEr, and BERTScore-F1~\cite{papineni2002bleu,banerjee2005meteor,lin2004rouge,vedantam2015cider,zhang2019bertscore}, following prior 3D-QA works. 
Evaluation is conducted using COCO Captions and Hugging Face Evaluate.
All experiments follow the protocol of 3D-LLM~\cite{hong20233dllm}. 
For methods without public implementations, we report results from the original papers.

\subsection{Main Results on ScanQA}
\label{sec:exp_main_scanqa}

Table~\ref{exp_scanqa_val_main_results} presents results on the ScanQA validation set. 
Despite being fully training-free, ViewMind3D achieves competitive performance compared to both 3D-based and fine-tuned VLM baselines. 
ViewMind3D (GPT-4.1) attains results comparable to 3D-LLM across multiple metrics, particularly on METEOR, demonstrating that structured zero-shot reasoning can approach the performance of trained 3D models. 
ViewMind3D (OpenAI o3) further improves semantic metrics, achieving the best performance on multiple metrics, including BLEU-1, METEOR, CIDEr, and BERTScore-F1.

Recent approaches such as GPT4Scene (GPT-4o, zero-shot) report performance on the ROUGE-L metric, benefiting from multi-view visual prompting and enriched spatial context through BEV representations and spatial-temporal markers. 
While effective, these gains rely on carefully designed input representations. 
In contrast, our method achieves competitive performance through structured reasoning and modular orchestration, without requiring specialized input design or additional adaptation.

We observe that GPT-4.1-MultiView achieves lower BLEU-1 but higher BERTScore-F1 than the SingleImage variant, indicating that multi-view inputs produce more descriptive outputs. 
While such outputs may reduce n-gram overlap, semantic metrics better capture their correctness.
Overall, these results demonstrate that modular orchestration of LLMs and VLMs enables strong 3D reasoning performance without any task-specific training.

\subsection{Main Results on SQA3D}
\label{sec:exp_main_sqa3d}

Table~\ref{tab:sqa3d_main_result} reports results on the SQA3D test set. 
Despite being fully training-free, ViewMind3D achieves 50.75\% accuracy, outperforming prior baselines under the same evaluation protocol. 
Recent approaches such as GPT4Scene (GPT-4o, zero-shot) report performance on SQA3D using the average EM-1 metric, achieving 44.8\% accuracy.

The largest gain is observed in ``What''-type questions, where ViewMind3D reaches 44.38\% accuracy, exceeding 3D-LLM by over 7 points. 
These questions require fine-grained spatial and relational reasoning~\cite{ma2022sqa3d}, indicating that the proposed design effectively captures spatially grounded information.
Performance on ``Which''-type questions is lower, as these resemble implicit multiple-choice tasks requiring explicit comparison, which is less naturally handled by generative zero-shot models. 
This suggests that incorporating candidate selection or ranking mechanisms could further improve performance.

Our method demonstrates consistent performance across multiple question types, highlighting the effectiveness of structured reasoning.
Overall, the results demonstrate that structured decomposition enables general-purpose VLMs to achieve strong zero-shot performance on complex 3D-QA tasks without task-specific training.

\subsection{Efficiency and Computational Analysis}

We analyze the computational characteristics of ViewMind3D from a pipeline perspective. 
As a modular and training-free framework, the overall cost is primarily determined by the number of LLM/VLM invocations and the length of intermediate representations.

Based on empirical measurements, a full inference requires on average approximately 90K tokens and around 120 seconds, with moderate variance across queries. 
Among all components, the 3D-QA module constitutes the primary computational bottleneck, consuming the largest portion of tokens (approximately 35K on average) due to multi-view captioning, cross-view summarization, and answer synthesis. 
The Guided Detection module introduces additional overhead (around 30K tokens), mainly for language-conditioned grounding, while the Relevance View Selection module contributes approximately 25K tokens.

This distribution indicates that computational cost is largely driven by downstream reasoning, while earlier stages help organize inputs for subsequent processing. 
In particular, Relevance View Selection reduces redundant views, which indirectly affects the overall computational load by limiting the amount of information passed to later stages.

While modern LLMs support large context windows, directly processing all multi-view observations in a single pass can be inefficient and may complicate reasoning over irrelevant inputs. 
In contrast, the proposed pipeline decomposes the 3D-QA task into multiple stages, distributing computation across components.

This modular design provides flexibility in balancing performance and efficiency. 
For example, the number of selected views can be adjusted, or certain modules can be simplified depending on task requirements, without requiring retraining.

\begin{table*}[t]
\centering
\small
\setlength{\tabcolsep}{5pt}
\renewcommand{\arraystretch}{1}

\caption{
Component-wise ablation results on ScanQA 
(VI=Viewpoint Indicator, GD=Guided Detection, RD=Role Decomposition, TBI=Term-Based Instructions).
}
\begin{tabular}{lcccccc}
\toprule
 & BLEU-1 $\uparrow$ & METEOR $\uparrow$ & ROUGE-L $\uparrow$ & CIDEr $\uparrow$ & Cos. Sim. $\uparrow$ & BERTScore-F1 $\uparrow$ \\
\midrule

ViewMind3D w/o VI     & \textbf{29.72} & 14.32 & 33.60 & 57.41 & 60.28 & \textbf{88.04} \\
ViewMind3D w/ VI      & 29.32 & \textbf{15.02} & \textbf{35.22} & \textbf{63.48} & \textbf{60.29} & 87.70 \\
\midrule

ViewMind3D w/o GD     & 27.65 & 14.99 & 34.40 & 60.30 & 59.35 & 86.27 \\
ViewMind3D w/ GD      & \textbf{29.58} & \textbf{15.34} & \textbf{35.38} & \textbf{61.45} & \textbf{60.04} & \textbf{87.86} \\
\midrule

ViewMind3D w/o RD     & 27.02 & \textbf{13.89} & 31.40 & 54.77 & 57.10 & 86.09 \\
ViewMind3D w/ RD      & \textbf{28.01} & 13.62 & \textbf{32.03} & \textbf{58.05} & \textbf{58.65} & \textbf{87.56} \\
\midrule

ViewMind3D w/o TBI     & 28.01 & 13.62 & 32.03 & 58.05 & 58.65 & 87.56 \\
ViewMind3D w/ TBI      & \textbf{29.32} & \textbf{15.02} & \textbf{35.22} & \textbf{63.48} & \textbf{60.29} & \textbf{87.70} \\
\bottomrule
\end{tabular}

\vspace{3pt}
\footnotesize{
\textbf{Note:} Each ablation pair is evaluated under a component-specific setting. In particular, GDM uses a subset for efficiency, and role decomposition and TBI are tested under different prompt configurations. Comparisons are valid only within each pair.
}

\label{tab: abl_scanqa}
\end{table*}

\begin{table*}[t]
\centering
\small
\setlength{\tabcolsep}{6pt}
\renewcommand{\arraystretch}{1}

\caption{
Component-wise ablation results on SQA3D.
}
\begin{tabular}{lccccccc}
\toprule
 & What & Is & How & Can & Which & Others & Avg. \\
\midrule

ViewMind3D w/o VI & 38.97 & 61.20 & \textbf{50.75} & 60.06 & 30.20 & 49.29 & 47.46 \\
ViewMind3D w/ VI  & \textbf{39.49} & \textbf{63.80} & 47.53 & \textbf{60.95} & \textbf{37.89} & \textbf{50.00} & \textbf{48.65} \\
\midrule

ViewMind3D w/o GD & 38.86 & 63.08 & \textbf{47.31} & \textbf{61.19} & 24.29 & 50.91 & 47.07 \\
ViewMind3D w/ GD  & \textbf{39.74} & \textbf{64.62} & 39.78 & 59.70 & \textbf{40.00} & \textbf{54.55} & \textbf{48.64} \\
\bottomrule
\end{tabular}

\label{tab: abl_sqa3d}
\end{table*}
% =============

\section{Ablation and Analysis}

\subsection{Component-wise Ablation}

We remove two specific modules, Guided Detection and Viewpoint Indicator, in ViewMind3D to validate each of their contributions.
All experiments use the same GPT-4.1 backbone with identical prompt settings. No rule-based post-processing is conducted to ensure consistent conditions. For SQA3D, we adopt a mixed prompt strategy: a compass-based protocol for ``What'' and ``Which'' questions, and a clock-based protocol for other question types, based on empirical observations.

\textbf{Effectiveness of Guided Detection.}
Incorporating this module improves performance across several metrics on ScanQA, with CIDEr increasing from 60.30 to 61.45 (Table~\ref{tab: abl_scanqa}), indicating that object-centric grounding enhances alignment between visual evidence and generated answers. 
A similar trend is observed on SQA3D, where the average accuracy improves from 47.07 to 48.64 (Table~\ref{tab: abl_sqa3d}).

However, the improvements are not uniform across question types. In particular, performance on ``How'' questions decreases (47.31 $\rightarrow$ 39.78, Table~\ref{tab: abl_sqa3d}). This behavior reflects a limitation of open-world object detectors, which may introduce noisy or inaccurate annotations for unseen or ambiguous categories. 
Such noise is especially detrimental to count-based or procedural reasoning, where precise and reliable visual cues are required.

These findings suggest that object-centric grounding is beneficial when accurate detection is available, but may become less effective when the grounding signal is noisy or misaligned with the reasoning objective. 
To address this issue, we can adaptively bypass Guided Detection for ``How'' questions by directly using raw image inputs instead.

\textbf{Effectiveness of Viewpoint Indicator.}
Enabling this module leads to a noticeable improvement in CIDEr on ScanQA (57.41 $\rightarrow$ 63.48, Table~\ref{tab: abl_scanqa}), with additional gains in METEOR and ROUGE-L, indicating that explicit camera pose cues support spatial grounding and content alignment. On SQA3D, we see improvements across most question types, with the largest gain observed for ``Which'' questions (30.20 $\rightarrow$ 37.89, Table~\ref{tab: abl_sqa3d}). 
This suggests that viewpoint-aware representations are particularly useful for resolving reference-based queries that require spatial disambiguation among multiple candidates.

The relatively modest gains on other metrics can be attributed to three factors. 
First, the input views are sampled from the $\mathrm{scannet\_frames\_25k}$ subset, resulting in sparse multi-view coverage that limits the impact of camera pose cues. Second, Relevance View Selection already reduces redundant viewpoints, constraining multi-view evidence aggregation. Third, View Captioner already incorporates semantic viewpoint descriptors, partially compensating for the absence of explicit geometric indicators.

Overall, these results indicate that Viewpoint Indicator primarily benefits tasks requiring explicit spatial disambiguation, while its contribution is less pronounced when semantic cues alone are sufficient.

\subsection{Single-Agent vs. Multi-Agent Ablation}

We compare the standard ViewMind3D pipeline, which assigns multiple specialized agents for question interpretation, view captioning, and cross-view summarization, against a collapsed baseline where all functionalities are handled by a single agent. We call the former \textit{role decomposition}. For consistency, Term-Based Instructions is disabled in both settings.

As shown in Table~\ref{tab: abl_scanqa}, role decomposition improves performance across multiple metrics (CIDEr 54.77 $\rightarrow$ 58.05). 
Similar gains are observed in BLEU-1, ROUGE-L, cosine similarity, and BERTScore-F1, suggesting that structured, modular reasoning leads to more semantically aligned answers.
In contrast, the monolithic baseline must simultaneously handle all reasoning tasks, which may introduce ambiguity and instability.

These findings indicate that the performance gains are primarily attributable to the modular pipeline design of ViewMind3D, rather than prompt variation alone. 
Such role decomposition structures the reasoning process through intermediate stages without requiring additional training.

\subsection{Ablation on Term-Based Instructions}

Using the Term-Based Instructions setting, our agents are constrained to use a canonical vocabulary for object categories, colors, and spatial descriptors from frequently occurring terms in the human-annotated answers in ScanQA. As shown in Table~\ref{tab: abl_scanqa}, this setting improves n-gram-based metrics (CIDEr 58.05 $\rightarrow$ 63.48), as such metrics are sensitive to lexical overlap and exact phrasing.

Without such setting, large language models tend to generate diverse and semantically valid expressions, including synonyms and paraphrases, which may deviate from the canonical vocabulary used in ground truth annotations. 
While these variations are linguistically appropriate, they can reduce surface-level overlap and lead to lower scores under string-based evaluation metrics.

These findings suggest that Term-Based Instructions function as an output normalization mechanism that aligns model predictions with benchmark evaluation protocols. 
Rather than altering the reasoning capability, it reduces discrepancies between semantically correct answers and their canonical representations. From a system perspective, this module complements other components in ViewMind3D by stabilizing the final output space.

\begin{table*}[ht]
\centering
\caption{Ablation results on BEV direction encoding schemes.}
\setlength{\tabcolsep}{4pt}
\begin{tabular}{lccccccc}
\toprule
\textbf{ScanQA (subset)} & BLEU-1 $\uparrow$ & METEOR $\uparrow$ & ROUGE-L $\uparrow$ & CIDEr $\uparrow$ & Cos. Sim. $\uparrow$ & BERTScore-F1 $\uparrow$ \\
\midrule
ViewMind3D (Compass) & \textbf{29.58} & 15.34 & 35.38 & \textbf{61.45} & 60.04 & 87.86 \\
ViewMind3D (Clock)   & 29.11 & \textbf{15.46} & \textbf{35.77} & 60.89 & \textbf{61.26} & \textbf{89.24} \\
\midrule

\multicolumn{7}{l}{} \\
\midrule
\textbf{SQA3D} & What & Is & How & Can & Which & Others & Avg. \\
\midrule
ViewMind3D (Compass) & 43.16 & \textbf{63.96} & \textbf{40.86} & 58.58 & \textbf{37.89} & 50.53 & 48.85 \\
ViewMind3D (Clock)   & \textbf{43.50} & 62.27 & 40.43 & \textbf{59.47} & 37.61 & \textbf{53.71} & \textbf{49.16} \\
\bottomrule
\label{tab:abl_directional}
\end{tabular}

\vspace{2pt}
\footnotesize{
\textbf{Note:} Both encoding schemes yield comparable performance. Clock-based encoding provides finer spatial resolution, benefiting detailed reasoning, while compass-based encoding remains effective for coarse directional cues.}
\end{table*}

\subsection{Ablation on BEV-Based Direction Encoding}

We compare two BEV direction encoding schemes, compass-based and clock-based, on ScanQA and SQA3D. 
As shown in Table~\ref{tab:abl_directional}, both schemes achieve comparable performance across all metrics, indicating that ViewMind3D is not sensitive to the choice of directional encoding.

Since SQA3D includes diverse question types, we observe moderate performance variations across them. 
The clock-based scheme performs better on ``What'' and ``Others'' questions, which often require finer spatial discrimination, while the compass-based scheme shows advantages on ``Is'' questions, where coarse directional reasoning is typically sufficient.

These results suggest that the required level of spatial granularity depends on the question. 
Clock-based encoding provides finer angular resolution for detailed spatial reasoning, whereas compass-based encoding offers simpler directional cues for coarse decisions. 
An adaptive hybrid approach can select between compass- and clock-based encoding based on the question type.

\section{Conclusion}

We presented \textbf{ViewMind3D}, a training-free and modular framework for 3D question answering in complex indoor scenes. By decomposing 3D-QA into VLM-driven components—view selection, visual grounding, viewpoint indication, and structured answer synthesis—the framework enables multi-modal reasoning without 3D-specific training. Experiments on ScanQA and SQA3D demonstrate competitive performance against fine-tuned 3D-LLMs.

Beyond performance, ViewMind3D offers strong interpretability through explicit intermediate representations and supports flexible integration with diverse LLM/VLM backbones. Overall, our results highlight that structured decomposition is an effective and practical approach for training-free 3D vision-language reasoning.

\bibliographystyle{IEEEtran}
\bibliography{ref} 

@INPROCEEDINGS{DSPNet,
  author={Luo, Jingzhou and Liu, Yang and Chen, Weixing and Li, Zhen and Wang, Yaowei and Li, Guanbin and Lin, Liang},
  booktitle={IEEE/CVF Conf. on Computer Vision and Pattern Recognition}, 
  title="{DSPNet: Dual-vision Scene Perception for Robust 3D Question Answering}", 
  year={2025}}

@article{MVScanQA,
  title="{Advancing 3D Scene Understanding with {MV-ScanQA} Multi-View Reasoning Evaluation and TripAlign Pre-training Dataset}",
  author={Wentao Mo and Qingchao Chen and Yuxin Peng and Siyuan Huang and Yang Liu},
  journal={ACM International Conference on Multimedia},
  year={2025}
}

@inproceedings{Liu2023VisualIT,
  author    = {Haotian Liu and Chunyuan Li and Qingyang Wu and Yong Jae Lee},
  title     = {{Visual Instruction Tuning}},
  booktitle = {Advances in Neural Information Processing Systems (NeurIPS)},
  year      = {2023},
  volume    = {36}
}

@misc{Achiam2023GPT4TR,
  author = {OpenAI},
  title  = {{GPT-4}},
  year   = {2023},
  url    = {https://openai.com/index/gpt-4-research/},
  note   = {Accessed: 2025-08-14}
}

@misc{openai2023gpt4v,
  author = {OpenAI},
  title  = {{GPT-4V(ision) System Card}},
  year   = {2023},
  url    = {https://openai.com/index/gpt-4v-system-card},
  note   = {Accessed: 2025-08-14}
}

@inproceedings{chen2020scanrefer,
  author    = {Dave Zhenyu Chen and Angel X. Chang and Matthias Nie{\ss}ner},
  title     = {{ScanRefer: 3D Object Localization in RGB-D Scans Using Natural Language}},
  booktitle = {European Conference on Computer Vision (ECCV)},
  year      = {2020}
}

@inproceedings{azuma2022scanqa,
  author    = {Daichi Azuma and Taiki Miyanishi and Shuhei Kurita and Motoaki Kawanabe},
  title     = {{ScanQA: 3D Question Answering for Spatial Scene Understanding}},
  booktitle = {IEEE/CVF Conference on Computer Vision and Pattern Recognition (CVPR)},
  year      = {2022}
}

@inproceedings{ma2022sqa3d,
  author    = {Xiaojian Ma and Silong Yong and Zilong Zheng and Qing Li and Yitao Liang and Song-Chun Zhu and Siyuan Huang},
  title     = {{SQA3D: Situated Question Answering in 3D Scenes}},
  booktitle = {International Conference on Learning Representations (ICLR)},
  year      = {2023}
}

@inproceedings{hong20233dllm,
  author        = {Hong, Yining and Zhen, Haoyu and Chen, Peihao and Zheng, Shuhong and Du, Yilun and Chen, Zhenfang and Gan, Chuang},
  title         = {{3D-LLM: Injecting the 3D World into Large Language Models}},
  booktitle     = {Advances in Neural Information Processing Systems},
  year          = {2023},
  volume        = {36}
}

@inproceedings{Radford2021CLIP,
  author    = {Alec Radford and Jong Wook Kim and Chris Hallacy and Aditya Ramesh and Gabriel Goh and Sandhini Agarwal and others},
  title     = {{Learning Transferable Visual Models From Natural Language Supervision}},
  booktitle = {International Conference on Machine Learning (ICML)},
  year      = {2021}
}

@inproceedings{Li2022BLIPBL,
  author    = {Junnan Li and Dongxu Li and Caiming Xiong and Steven C. H. Hoi},
  title     = {{BLIP: Bootstrapping Language-Image Pre-training for Unified Vision-Language Understanding and Generation}},
  booktitle = {International Conference on Machine Learning (ICML)},
  year      = {2022}
}

@inproceedings{xu2024pointllm,
  author    = {Runsen Xu and Xiaolong Wang and Tai Wang and Yilun Chen and Jiangmiao Pang and Dahua Lin},
  title     = {{PointLLM: Empowering Large Language Models to Understand Point Clouds}},
  booktitle = {European Conference on Computer Vision (ECCV)},
  year      = {2024}
}

@inproceedings{xiao2024florence2,
  author    = {Bin Xiao and Haiping Wu and Weijian Xu and Xiyang Dai and Houdong Hu and Yumao Lu and others},
  title     = {{Florence-2: Advancing a Unified Representation for a Variety of Vision Tasks}},
  booktitle = {IEEE/CVF Conference on Computer Vision and Pattern Recognition (CVPR)},
  year      = {2024}
}

@inproceedings{qi2024shapellm,
  author    = {Zekun Qi and Runpei Dong and Shaochen Zhang and Haoran Geng and Chunrui Han and Zheng Ge and others},
  title     = {{ShapeLLM: Universal 3D Object Understanding for Embodied Interaction}},
  booktitle = {European Conference on Computer Vision (ECCV)},
  year      = {2024}
}

@article{chen2024grounded,
  author        = {Yilun Chen and Shuai Yang and Haifeng Huang and Tai Wang and Runsen Xu and Ruiyuan Lyu and others},
  title         = {{Grounded 3D-LLM with Referent Tokens}},
  journal       = {arXiv preprint arXiv:2405.10370},
  eprint        = {2405.10370},
  archivePrefix = {arXiv},
  year          = {2024}
}

@inproceedings{chen2024ll3da,
  author    = {Sijin Chen and Xin Chen and Chi Zhang and Mingsheng Li and Gang Yu and Hao Fei and others},
  title     = {{LL3DA: Visual Interactive Instruction Tuning for Omni-3D Understanding Reasoning and Planning}},
  booktitle = {IEEE/CVF Conference on Computer Vision and Pattern Recognition (CVPR)},
  year      = {2024}
}

@inproceedings{yang2024llmgrounder,
  author    = {Jianing Yang and Xuweiyi Chen and Shengyi Qian and Nikhil Madaan and Madhavan Iyengar and David F. Fouhey and Joyce Chai},
  title     = {{LLM-Grounder: Open-Vocabulary 3D Visual Grounding With Large Language Model as an Agent}},
  booktitle = {IEEE International Conference on Robotics and Automation (ICRA)},
  year      = {2024}
}

@inproceedings{dai2017scannet,
  author    = {Angela Dai and Angel X. Chang and Manolis Savva and Maciej Halber and Thomas Funkhouser and Matthias Nie{\ss}ner},
  title     = {{ScanNet: Richly-Annotated 3D Reconstructions of Indoor Scenes}},
  booktitle = {IEEE Conference on Computer Vision and Pattern Recognition (CVPR)},
  year      = {2017}
}

@inproceedings{Li2023BLIP2BL,
  author    = {Junnan Li and Dongxu Li and Silvio Savarese and Steven C. H. Hoi},
  title     = {{BLIP-2: Bootstrapping Language-Image Pre-training with Frozen Image Encoders and Large Language Models}},
  booktitle = {International Conference on Machine Learning (ICML)},
  year      = {2023}
}

@inproceedings{alayrac2022flamingo,
  author    = {Jean-Baptiste Alayrac and Jeff Donahue and Pauline Luc and Antoine Miech and Iain Barr and Yana Hasson and others},
  title     = {{Flamingo: A Visual Language Model for Few-Shot Learning}},
  booktitle = {Advances in Neural Information Processing Systems (NeurIPS)},
  year      = {2022}
}

@inproceedings{papineni2002bleu,
  author    = {Kishore Papineni and Salim Roukos and Todd Ward and Wei-Jing Zhu},
  title     = {{BLEU: A Method for Automatic Evaluation of Machine Translation}},
  booktitle = {Annual Meeting of the Association for Computational Linguistics (ACL)},
  year      = {2002}
}

@inproceedings{banerjee2005meteor,
  author    = {Satanjeev Banerjee and Alon Lavie},
  title     = {{METEOR: An Automatic Metric for MT Evaluation with Improved Correlation with Human Judgments}},
  booktitle = {ACL Workshop on Intrinsic and Extrinsic Evaluation Measures for Machine Translation and/or Summarization},
  year      = {2005}
}

@inproceedings{lin2004rouge,
  author    = {Chin-Yew Lin},
  title     = {{ROUGE: A Package for Automatic Evaluation of Summaries}},
  booktitle = {ACL Workshop on Text Summarization Branches Out},
  year      = {2004}
}

@inproceedings{vedantam2015cider,
  author    = {Ramakrishna Vedantam and C. Lawrence Zitnick and Devi Parikh},
  title     = {{CIDEr: Consensus-Based Image Description Evaluation}},
  booktitle = {IEEE Conference on Computer Vision and Pattern Recognition (CVPR)},
  year      = {2015}
}

@article{zhang2019bertscore,
  author        = {Tianyi Zhang and Varsha Kishore and Felix Wu and Kilian Q. Weinberger and Yoav Artzi},
  title         = {{BERTScore: Evaluating Text Generation with BERT}},
  journal       = {arXiv preprint arXiv:1904.09675},
  eprint        = {1904.09675},
  archivePrefix = {arXiv},
  year          = {2019}
}

@INPROCEEDINGS{SceneLLM,
author={Fu, Rao and Liu, Jingyu and Chen, Xilun and Nie, Yixin and Xiong, Wenhan},
booktitle={IEEE/CVF Winter Conference on Applications of Computer Vision (WACV)},
title="{Scene-LLM: Extending Language Model for 3D Visual Reasoning}",
year={2025}
}

@article{GPT4Scene,
title="{GPT4Scene: Understand 3D Scenes from Videos with Vision-Language Models}",
author={Zhangyang Qi and Zhixiong Zhang and Ye Fang and Jiaqi Wang and Hengshuang Zhao},
journal={arXiv preprint arXiv:2501.01428},
year={2024}
}

@article{ChatScene++,
  title="{Chat-Scene++: Exploiting Context-Rich Object Identification for 3D LLM}",
  author={Huang, Haifeng and Chen, Yilun and Wang, Zehan and Pang, Jiangmiao and Zhao, Zhou},
  journal={arXiv preprint arXiv:2603.27507},
  year={2026}
}

@article{ChatScene,
  title="{Chat-scene: Bridging 3d scene and large language models with object identifiers}",
  author={Huang, Haifeng and Chen, Yilun and Wang, Zehan and Huang, Rongjie and Xu, Runsen and Wang, Tai and Liu, Luping and Cheng, Xize and Zhao, Yang and Pang, Jiangmiao and others},
  journal={Advances in Neural Information Processing Systems (NeurIPS)},
  year={2024}
}

\end{document}